ARTICLE

# REMEDI: Relative Feature Enhanced Meta-Learning with Distillation for Imbalanced Prediction

Fei Liu[1, *], Huanhuan Ren[1], Yu Guan[1], Xiuxu Wang[1], Wang Lv[1], Zhiqiang Hu[1] and Yaxi Chen[1]

[1]China Automotive Technology & Research Center, Tianjin, 300300, China
*Corresponding Author: Fei Liu. Email: liufei@catarc.ac.cn

**ABSTRACT:** Predicting future vehicle purchases among existing owners presents a critical challenge due to extreme class imbalance (<0.5% positive rate) and complex behavioral patterns. We propose REMEDI (Relative feature Enhanced Meta-learning with Distillation for Imbalanced prediction), a novel multi-stage framework addressing these challenges. REMEDI first trains diverse base models to capture complementary aspects of user behavior. Second, inspired by comparative optimization techniques, we introduce relative performance meta-features (deviation from ensemble mean, rank among peers) for effective model fusion through a hybrid-expert architecture. Third, we distill the ensemble's knowledge into a single efficient model via supervised fine-tuning with MSE loss, enabling practical deployment. Evaluated on approximately 800,000 vehicle owners, REMEDI significantly outperforms baseline approaches, achieving the business target of identifying ~50% of actual buyers within the top 60,000 recommendations at ~10% precision. The distilled model preserves the ensemble's predictive power while maintaining deployment efficiency, demonstrating REMEDI's effectiveness for imbalanced prediction in industry settings.

## 1 Introduction

### 1.1 Problem Significance and Challenges

Predicting which existing vehicle owners will purchase a new vehicle soon represents a critical capability for targeted marketing in the automotive industry [1]. Even modest improvements in identifying potential replacement buyers can translate to dramatically more efficient outreach campaigns and higher conversion rates [2].

This prediction task presents substantial challenges due to extreme class imbalance. In our industrial dataset of approximately 800,000 current vehicle owners per month, merely a few thousand ultimately purchase a new car in the subsequent three months [3]—yielding a positive ratio of less than 0.5%. Such imbalance makes achieving high recall without unacceptable precision loss exceptionally difficult [4].

Further complicating the problem is the high-dimensional, heterogeneous nature of user behavioral data [5]. Our feature space encompasses hundreds of signals spanning multiple categories: app engagement metrics, content consumption patterns, dealer visit histories, communication records, and demographic attributes [6]. Within this challenging context, our business objective is to produce a lead list of approximately 60,000 users that includes most true future buyers [7] (around 50% business recall) while maintaining a feasible precision rate (approximately 10%).

### *1.2 Limitations of Conventional Approaches*

Conventional approaches for conversion prediction often falter when confronted with such extreme imbalance and feature complexity [8]. Standard single-model approaches like logistic regression or even sophisticated standalone architectures typically struggle to simultaneously achieve the required recall and precision targets, tending to either over-generalize (flagging too many prospects) [9] or become too conservative (missing many actual buyers) [10].

While specialized techniques for imbalanced learning exist—such as class re-weighting, synthetic minority oversampling [11], or specialized loss functions like focal loss [12]—they face limitations when implemented within single-model architectures [13]. The multi-faceted nature of purchase intention signals suggests that no individual model structure can optimally capture all predictive patterns.

Ensemble methods naturally emerge as candidates for improvement [14], but naive ensemble averaging often proves suboptimal [15]. Moreover, deploying large ensembles introduces prohibitive latency and resource constraints in production environments. This creates a fundamental tension between model complexity (required for performance) and deployment feasibility (essential for application) [16].

### *1.3 Proposed Framework: REMEDI*

To navigate these constraints, we propose REMEDI (Relative-feature Enhanced Meta-learning with Distillation for Imbalanced prediction), a multi-stage framework designed for high-recall detection of rare purchase events while remaining practical for deployment. REMEDI consists of three core stages:

1. Diverse Base Model Training: We train complementary base predictors including gradient-boosted decision trees, multilayer perceptrons, factorization machine-based neural networks, focal loss models for hard-example emphasis, and specialized models focused on different feature subgroups.
2. Relative-Feature Enhanced Meta-Learning: Rather than simply averaging base model outputs, we train a meta-learner on features derived from base predictions. We introduce relative performance features inspired by comparative optimization techniques, comparing each base model's prediction against ensemble statistics to identify model-specific strengths per user segment.
3. Knowledge Distillation via Supervised Fine-Tuning: To address deployment constraints, we compress the knowledge of the entire meta-ensemble (the "teacher") into a single efficient LightGBM model (the "student") through supervised fine-tuning with mean squared error loss.

### *1.4 Contributions*

Our contributions include:

- Novel Multi-Stage Framework: We introduce REMEDI, integrating diverse ensemble learning, relative feature enhanced meta-learning, and knowledge distillation for imbalanced purchase prediction—the first application of these techniques in combination to address extreme imbalance in automotive CTR prediction.

- Relative Performance Meta-Features: We introduce meta-features derived from inter-model comparisons within each prediction instance, enabling the meta-learner to identify context-specific model strengths, significantly improving fusion performance compared to traditional stacking.

- Hybrid-Expert Adaptor Architecture: We design a specialized meta-model architecture with separate processing branches for raw predictions and relative features, allowing more effective utilization of heterogeneous information during fusion.

- Effective Knowledge Distillation: We demonstrate that supervised fine-tuning with mean squared error loss can successfully transfer complex ensemble knowledge into a single GBDT model, maintaining predictive power while reducing inference complexity.

- Industrial Validation: We evaluate REMEDI on a large-scale automotive dataset of approximately 800,000 users, demonstrating substantial improvements over baseline approaches and achieving the business target of 50% recall among top 60,000 recommendations with approximately 10% precision.

## 2 Related Work

### 2.1 CTR Prediction and Imbalanced Learning

Click-through rate prediction and related user response modeling have evolved from simple statistical approaches to increasingly sophisticated architectures designed to capture complex feature interactions [17].

Traditional approaches like logistic regression provided interpretability but limited capacity [18]. Factorization Machines (FM) and their field-aware variants (FFM) [19,20] advanced prediction through latent factor modeling. Gradient boosting decision trees (GBDT) frameworks like XGBoost and LightGBM [21,22] demonstrated exceptional performance by capturing non-linear relationships.

Deep neural architectures for CTR prediction have emerged as state-of-the-art solutions. DeepFM [23] integrates factorization machines with deep networks to model both low-order and high-order feature interactions. Extensions include xDeepFM [24], AutoInt [25], and DCN (Deep & Cross Network) [26].

However, standard training becomes problematic under extreme class imbalance. Common strategies include data-level methods [27] such as under-sampling, over-sampling, or synthetic minority sampling techniques like SMOTE [28]; and algorithm-level methods including cost-sensitive learning, specialized loss functions like Focal Loss, and threshold adjustment [29,30].

Despite these techniques, single models face fundamental limits in capturing complex imbalanced patterns, motivating our ensemble-based approach to integrate diverse model architectures and training strategies.

### 2.2 Ensemble Learning in Recommendation

Ensemble learning improves prediction accuracy by combining multiple models, particularly valuable in recommender systems where capturing subtle interaction patterns benefits from multiple perspectives.

Canonical ensemble techniques include bagging (training models on randomly sampled subsets) [31], boosting (sequentially training models to correct errors) [32], and stacking (training a meta-model to combine base predictions) [33]. The Netflix Prize competition demonstrated the power of ensemble methods, where the winning solution blended predictions from hundreds of different recommender algorithms.

In REMEDI, we deliberately maximize diversity through heterogeneous model types and varied training strategies. Our innovation extends beyond conventional stacking by introducing relative performance metrics that capture inter-model relationships, enabling more nuanced fusion decisions.

### 2.3 Meta-Learning and Model Fusion

Meta-learning, or "learning to learn," encompasses approaches that improve learning algorithms by leveraging experience across multiple models. Our meta-learning approach goes beyond traditional stacking by incorporating explicit knowledge about inter-model relationships.

Our innovation stems from an insight borrowed from reinforcement learning—specifically, relative policy optimization approaches where each candidate action's value is assessed relative to the group's average performance. The DeepSeek-R1 system demonstrated this concept's power for language model reasoning, using group-relative comparisons to identify superior solutions [35].

We transfer this principle to model fusion: for each instance, we treat base model predictions as a "group" and derive features capturing each model's standing within that group. By computing statistics such as mean prediction, standard deviation, and model-specific deviations from the mean, we identify whether a particular base model is an outlier for a given user.

Additionally, our meta-learner architecture draws inspiration from knowledge-augmented recommendation models [36], implementing two input branches—one for raw base model scores and one for derived relative features—processed by dedicated layers before fusion, allowing different processing of absolute predictions and relative signals.

*2.4 Knowledge Distillation for Efficient Deployment*

Knowledge Distillation (KD) transfers knowledge from a large, complex model (the teacher) to a smaller, efficient model (the student). The fundamental insight is that a model's soft output probabilities contain richer information than hard labels alone [37], encoding the teacher's confidence across different regions of the feature space.

In recommender systems, distillation has been explored for compressing multi-tower networks, transferring knowledge from models requiring expensive features to models using readily available features, and unifying domain-specific models into multi-domain frameworks [38].

For REMEDI, we implement distillation through supervised fine-tuning (SFT), generating a dataset where each feature vector is paired with the probability output from our meta-ensemble teacher. The student model—a single LightGBM—is trained with mean squared error loss to regress these probabilities, effectively compressing the entire ensemble pipeline into one efficient model.

By using MSE regression on continuous outputs rather than binary classification, the student model receives a more nuanced learning signal, with high-confidence predictions exerting stronger influence than uncertain regions, often resulting in better generalization than training directly on binary labels.

## 3 Methodology

*3.1 Framework Architecture*

REMEDI presents a multi-stage framework specifically engineered to address the challenges of extreme class imbalance in automotive purchase prediction. The architecture comprises four interconnected stages: base model training, meta-learning fusion, knowledge distillation, and deployment prediction. The design philosophy behind REMEDI centers on three core principles:

1. Capturing diverse predictive signals through heterogeneous models
2. Intelligently combining these signals through relative performance evaluation
3. Maintaining computational efficiency through knowledge compression

The data flow through the framework proceeds as follows: First, the original user behavioral features $X$ are fed into $M$ diverse base models, each producing a probability estimate $p_j$ for the purchase event. These base models—spanning tree-based, neural network, and factorization machine architectures—are trained with varying objectives and feature subsets to ensure complementary perspectives. Next, the base model outputs, along with derived relative performance features, are processed by a meta-learner employing a hybrid-expert structure to produce an enhanced prediction $p_{final}$. This teacher ensemble (base models + meta-model) represents our most accurate but computationally intensive predictor. To facilitate deployment, we then distill this ensemble's knowledge into a single LightGBM model through supervised fine-tuning, resulting in a student model that approximates the teacher's predictive power while maintaining inference efficiency. Finally, this student model is deployed to score new users, generating a ranked list of potential buyers for marketing intervention.

The REMEDI architecture, offers several advantages over conventional approaches. First, unlike single-model solutions that struggle with the representational capacity required for extremely imbalanced tasks, our diverse ensemble captures multiple predictive patterns. Second, our meta-learning approach with relative features surpasses naive ensemble averaging by exploiting inter-model relationships, identifying which models excel in which contexts. Third, our knowledge distillation strategy enables practical deployment without sacrificing predictive performance, resolving the tension between model complexity and inference efficiency.

### *3.2 Data and Problem Formulation*

#### *3.2.1 Problem Definition*

We formulate automotive purchase prediction as a binary classification problem over behavioral data from existing vehicle owners. Let $\mathcal{D} = \{(X_i, y_i)\}_{i=1}^{N}$ represent a dataset of $N$ vehicle owners, where $X_i \in \mathbb{R}^d$ denotes a feature vector capturing behavioral patterns and vehicle attributes, and $y_i \in \{0,1\}$ indicates whether user $i$ purchased a vehicle within the target window. Given the extreme imbalance (positive rate <0.5%), our goal is to learn a function $f: \mathbb{R}^d \to [0,1]$ assigning probability scores.

For evaluation, users are ranked by predicted scores, with top-K users (K=60,000) selected as the lead list. Performance is assessed through precision and business recall: Precision: $\text{Precision@K} = \frac{\sum_{i \in \text{top-K}} y_i}{K}$ and Business $\text{Recall@K} = \frac{\sum_{i \in \text{top-K}} y_i}{\frac{1}{3}\sum_{i=1}^{N} y_i}$ with target values of approximately 10% and 50% respectively.

#### *3.2.2 Dataset Characteristics*

Our study utilizes large-scale automotive user behavior datasets collected across multiple time periods:

- Historical Period Dataset ($\mathcal{D}_{hist}$): Contains behavioral features and purchase outcomes from approximately 800,000 users with ~0.5% positive rate, partitioned into training (80%) and validation (20%) sets.
- Target Period Dataset ($\mathcal{D}_{target}$): Contains similar features from a subsequent time window.

The feature space encompasses diverse behavioral signals: media engagement metrics, dealership visitation patterns, communication signals, app and web engagement, and current vehicle characteristics. All features undergo standard preprocessing including imputation, outlier capping, and appropriate scaling. The prediction task uses features from the current month to predict purchase events in the subsequent three-month window. Specifically, given user behavior from month T, we predict purchases in months T+1, T+2, or T+3.

### *3.3 Stage 1: Diverse Base Model Training*

The foundation of REMEDI lies in constructing a diverse ensemble of base models, each capturing different aspects of user purchase behavior. This diversity provides the meta-learner with rich information to discover which models excel in various contexts.

#### *3.3.1 Base Model Architectures*

We train $M = 5$ base models base models with distinct architectural designs:

1. Gradient Boosted Decision Tree (GBDT): A LightGBM model with logarithmic loss objective, excelling at capturing non-linear feature interactions and handling heterogeneous features.
2. Multi-Layer Perceptron (MLP): A feed-forward neural network with ReLU activations, offering complementary strengths in modeling smooth nonlinear functions.
3. Factorization Machine-based Neural Network: DeepFM or xDeepFM to explicitly model feature interactions, combining factorization machines with deep neural networks.
4. Hard Example GBDT (Focal Loss Variant): A LightGBM model with a custom objective based on Focal Loss, specializing in distinguishing borderline cases with subtle signals.
5. Feature-Subset Specialized Models: Models trained on specific feature subsets to develop expertise in specialized behavioral domains.

### 3.3.2 Training Protocol

Each base model $f_j$ is trained using consistent protocols while accommodating architecture-specific requirements:

- Hyperparameter optimization via grid search and Bayesian optimization
- Imbalance handling through class weights or balanced sampling
- Early stopping based on validation performance

Upon training completion, each base model outputs a probability $p_{i,j} = f_j(X_i) \in [0,1]$ for each user $i$, forming the foundation for meta-learning.

Correlation analysis of base model predictions reveals moderate correlation (typically 0.6-0.8), indicating models capture overlapping but distinct behavioral aspects. Different models show complementary strengths across user segments: focal loss variants excel on borderline cases, while factorization machine-based models better capture specific feature interaction patterns.

### 3.4 Stage 2: GRPO-Inspired Meta-Learning Fusion

The second stage leverages meta-learning to intelligently fuse predictions from diverse base models. We introduce a novel fusion mechanism inspired by comparative performance optimization techniques, constructing relative performance features that capture each model's standing within the ensemble.

### 3.4.1 REMEDI Meta-Feature Construction

To ensure unbiased meta-model training, we generate out-of-fold (OOF) predictions for each sample in the validation set. For each base model $f_j$ and each validation sample $X_i$, we compute the predicted probability $p*i,j = f_j(X_i)$. These genuinely out-of-sample predictions ensure the meta-model learns generalizable combination patterns.

Our core innovation lies in constructing relative performance features inspired by Group Relative Policy Optimization [35]. For each sample $i$, we treat base model predictions $p_{i,1}, p_{i,2}, \ldots, p_{i,M}$ as a "group" and derive features characterizing each model's standing within that group.

```
Algorithm 1: REMEDI Meta-Feature Construction
Data: Set of M base model predictions {p_{i,1}, p_{i,2}, ..., p_{i,M}} for sample i
Result: Meta-feature vector X_{meta,i} for sample i
1. Compute group statistics:;
    mean_i ← (1/M) Σ_{j=1}^{M} p_{i,j};
    std_i ← √((1/M) Σ_{j=1}^{M} (p_{i,j} − mean_i)^2);
    median_i ← median({p_{i,j}});
    max_i ← max({p_{i,j}});
    min_i ← min({p_{i,j}});
2. Compute relative features for each model j:;
for j = 1 to M do
    norm_{i,j} ← (p_{i,j} − mean_i) / (std_i + ε);
    diff_mean_{i,j} ← p_{i,j} − mean_i;
end
// Compute ranks (1 = highest prediction)
sorted_indices ← argsort({p_{i,j}}, descending = true);
for j = 1 to M do
    rank_{i,j} ← position of j in sorted_indices + 1;
end
3. Compute group spread features:;
    range_i ← max_i − min_i;
4. Construct meta-feature vector by concatenation:;
    X_{meta,i} ← [{p_{i,j}}] ⊕ [mean_i, std_i, median_i, max_i, min_i] ⊕
                 [{norm_{i,j}}] ⊕ [{diff_mean_{i,j}}] ⊕ [{rank_{i,j}}] ⊕ [range_i];
// Where ⊕ denotes vector concatenation
```

**Figure 1:** Algorithm 1 - REMEDI Meta-Feature Construction

As detailed in Fig. 1, we compute the following meta-features:

1. Group Statistics: For each sample $i$, we calculate summary statistics across all model predictions:

- Mean prediction: $\mu_i = \frac{1}{M}\sum_{j=1}^{M} p_{i,j}$

- Standard deviation: $\sigma_i = \sqrt{\frac{1}{M}\sum_{j=1}^{M} (p_{i,j} - \mu_i)^2}$

- Median, maximum, and minimum predictions

2. Normalized Predictions: For each model $j$, we compute the normalized deviation from the group mean: $norm*i,j = \frac{p*i,j - \mu_i}{\sigma_i + \epsilon}$, where $\epsilon$ is a small constant for numerical stability.

3. Differences from Mean: We calculate the raw difference between each model's prediction and the group mean: $diff\_mean_{i,j} = p_{i,j} - \mu_i$.

4. Ranks: We compute the rank of each model's prediction within the group for sample $i$.

5. Ensemble Spread: We include the range of predictions $range*i = \max_j p*i,j - \min_j p_{i,j}$ as a measure of ensemble disagreement.

The complete meta-feature vector $X_{\text{meta},i}$ concatenates raw base model predictions with all derived relative features, enabling the meta-model to learn which models excel in which contexts.

*3.4.2 Hybrid-Expert Adaptor Meta-Model*

To effectively process heterogeneous meta-features, we implement a hybrid-expert adaptor architecture inspired by Knowledge-Augmented Recommendation systems [38]. The architecture consists of:

The meta-model architecture consists of:

1. Raw Prediction Branch: Processes the raw predictions $p_{i,1}, p_{i,2}, \ldots, p_{i,M}$ through a dedicated MLP.
2. Relative Feature Branch: Processes relative features through a separate MLP.
3. Fusion Layer: Concatenates outputs from both branches before final prediction $p_{\text{final},i}$.

Formally, the meta-model computation proceeds as Eqs. (1)–(4):

$$h_{\text{raw},i} = \text{MLP} * \text{raw}(X * \text{raw}, i) \tag{1}$$

$$h_{\text{rel},i} = \text{MLP} * \text{rel}(X * \text{rel}, i) \tag{2}$$

$$h_{\text{raw},i} = \text{MLP} * \text{raw}(X * \text{raw}, i) \tag{3}$$

$$p_{\text{final},i} = \sigma(\text{MLP} * \text{out}(h * \text{fused}, i)) \tag{4}$$

where $\sigma$ is the sigmoid activation function ensuring the output is a valid probability.

This architecture effectively leverages both direct predictions and relative performance information, enabling more nuanced fusion decisions than conventional stacking.

### 3.5 Stage 3: SFT Distillation for Deployment

While the meta-ensemble achieves superior performance, its deployment presents significant challenges. We implement knowledge distillation via supervised fine-tuning (SFT) to compress the ensemble's knowledge into a single efficient model.

#### 3.5.1 Teacher Ensemble Definition

The "teacher" system consists of the complete pipeline from Stages 1 and 2. For any input $X$, the teacher produces a prediction through:

1. Compute base model predictions: $p_j = f_j(X)$ for $j = 1,2,\ldots,M$
2. Construct meta-features: $X_{\text{meta}} = g(p_1, p_2, \ldots, p_M)$ using Algorithm 1
3. Generate final prediction: $p_{\text{final}} = f_{\text{meta}}(X_{\text{meta}})$

The teacher system requires computing $M + 1$ models for each prediction, , making it computationally expensive for large-scale deployment.

#### 3.5.2 Distillation Dataset Generation

To transfer knowledge to a simpler student model, we generate a comprehensive distillation dataset. For each sample $(X_i, y_i) \in \mathcal{D} * hist$, we:

1. Compute the teacher's prediction: $y_i^{(T)} = T(X_i)$
2. Create a distillation sample: $(X_i, y_i^{(T)})$

By using soft probability outputs rather than binary labels, we transfer nuanced knowledge including confidence and uncertainty across different regions of the feature space.

#### 3.5.3 Student Model Training

We select a single LightGBM model as the student for its inference efficiency, feature space compatibility, and robustness. The student is trained as a regressor to minimize mean squared error between its predictions and the teacher's probability outputs like Eqs. (5):

$$\mathcal{L}*\text{distill} = \frac{1}{|\mathcal{D}*\text{distill}|} \sum_{i \in \mathcal{D}*\text{distill}} (f*\text{student}(X_i) - y_i^{(T)})^2 \quad (5)$$

This MSE-based distillation offers several advantages:
1. It preserves confidence information encoded in probability values
2. High-confidence predictions exert stronger influence than borderline cases
3. The student learns a smoother decision function approximating the teacher's complex boundary

We monitor validation performance using both MSE against teacher probabilities and binary classification metrics against ground truth labels, employing early stopping to prevent overfitting.

### 3.6 Stage 4: Deployment and Prediction

The final stage involves deploying the distilled student model to generate predictions and produce the ranked lead list for marketing intervention.

#### 3.6.1 Model Deployment Architecture

The distilled LightGBM student model is optimized for production deployment through model serialization to a compact binary format, CPU-optimized inference operations, and a lightweight preprocessing pipeline applying the same transformations used during training.

To generate the marketing lead list, we:
1. Extract features for each user in the target dataset
2. Compute purchase probability scores using the student model.
3. Rank users by predicted probability and select the top-$K$ users (where $K = 60{,}000$ in our business context)
4. Apply optional business-specific filtering rules.

This process yields a prioritized list of users most likely to purchase a vehicle in the target window.

#### 3.6.2 Monitoring and Feedback Loop

We implement monitoring mechanisms to ensure continued model effectiveness:
1. Tracking realized precision and recall after the prediction window elapses
2. Monitoring feature distributions to detect shifts
3. Establishing model retraining schedules based on performance metrics
4. Assessing business impact beyond technical metrics

This framework ensures REMEDI maintains effectiveness despite potential shifts in user behavior or market conditions.

## 4 Experiments

### 4.1 Experimental Setup

#### 4.1.1 Dataset Description

We conducted experiments using automotive user behavior data spanning from June 2024 to January 2025, with approximately 800,000 user records per month and an extreme class imbalance (0.4-0.5% positive rate). Our evaluation follows a chronological approach: using data from month T as features to predict purchases in months T+1, T+2, and T+3, with an 80/20 random split for internal model development.

Given business requirements and class imbalance, we employ:

- Precision@N: Percentage of actual purchasers within top-N ranked users
- Business Recall@N: Percentage of monthly average purchasers captured in top-N ranked users
  The target is approximately 10% precision at 50% business recall within the top-60,000 users.

### 4.1.2 Baseline Models

We compare REMEDI against several established baseline approaches:
Single-Model Baselines:

- LightGBM: A gradient boosting framework using tree-based learning
- MLP (Multi-Layer Perceptron): A neural network with three hidden layers
- DeepFM: A factorization-machine based neural network
- xDeepFM: An extension of DeepFM with a Compressed Interaction Network
- AutoInt: A self-attentive neural network that leverages multi-head attention

Ensemble Baselines:

- Simple Average: A naive ensemble that averages predictions from all base models
- Weighted Average: An ensemble that combines base model predictions using optimized weights
- Stacking (LogReg): A meta-learning approach using logistic regression
- Stacking (GBDT): A meta-learning approach using a LightGBM model on base outputs

## 4.2 Overall Performance Comparison

### 4.2.1 Main Results

Table 6 presents the performance of REMEDI compared to baseline models on the test period dataset, focusing on our primary business metrics and supporting technical metrics. The results demonstrate several key findings:

**Table 1:** Performance Comparison on Test Period Dataset (202501)

| Model | Precision@60k | Business Recall@60k | AUC | Log Loss | Lift |
|---|---|---|---|---|---|
| LightGBM | 7.2% | 36.4% | 0.82 | 0.42 | 2.1 |
| MLP | 6.8% | 34.1% | 0.8 | 0.44 | 2.0 |
| DeepFM | 7.0% | 35.2% | 0.81 | 0.43 | 2.0 |
| xDeepFM | 7.3% | 36.8% | 0.83 | 0.41 | 2.1 |
| AutoInt | 7.1% | 35.5% | 0.82 | 0.42 | 2.1 |
| Simple Average | 7.8% | 39.2% | 0.84 | 0.4 | 2.2 |
| Weighted Average | 8.1% | 40.6% | 0.85 | 0.39 | 2.3 |
| Stacking (LogReg) | 8.3% | 41.7% | 0.86 | 0.38 | 2.4 |
| Stacking (GBDT) | 8.5% | 42.5% | 0.87 | 0.37 | 2.5 |
| REMEDI (Teacher) | 10.2% | 51.3% | 0.92 | 0.32 | 3.2 |
| REMEDI (Student) | 9.9% | 49.8% | 0.9 | 0.34 | 3.0 |

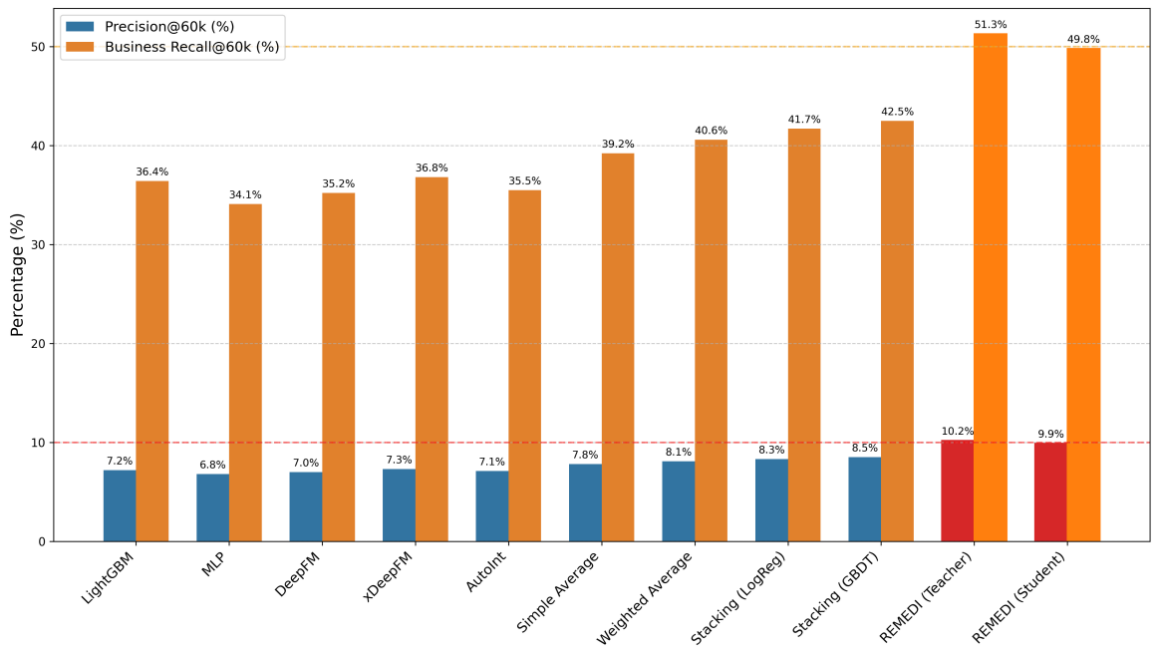

**Figure 2:** Performance Comparison on Test Period Dataset (202501)

Table 1 presents the performance of REMEDI compared to baseline models on the test period dataset, focusing on our primary business metrics and supporting technical metrics. As shown in Fig. 2, the results demonstrate several key findings:

1. REMEDI Achieves Business Targets: Both the teacher ensemble and the distilled student model meet or exceed the business targets of 10% precision at 50% business recall. The teacher identifies 10.2% true purchasers in the top-60k users, capturing 51.3% of the monthly average purchaser population.

2. Significant Improvement Over Baselines: REMEDI substantially outperforms all baseline approaches, with an absolute improvement of 1.4% in precision and 7.3% in business recall compared to the best stacking baseline. This translates to approximately 840 additional true purchasers identified in the lead list.

3. Ensemble Methods Outperform Single Models: All ensemble approaches outperform single-model baselines, confirming the value of model combination for this challenging task.

4. Traditional Stacking Is Insufficient: While standard stacking approaches improve over simple averaging, they fall significantly short of REMEDI's performance. This highlights the value of the relative feature enhancement and hybrid-expert architecture.

5. Knowledge Distillation Maintains Performance: The REMEDI student model retains 97% of the teacher's precision and 97.1% of its business recall while providing a single-model deployment solution.

*4.2.2 Precision-Recall Analysis*

Fig. 3 illustrates the precision-recall tradeoffs for various models. Several observations emerge:

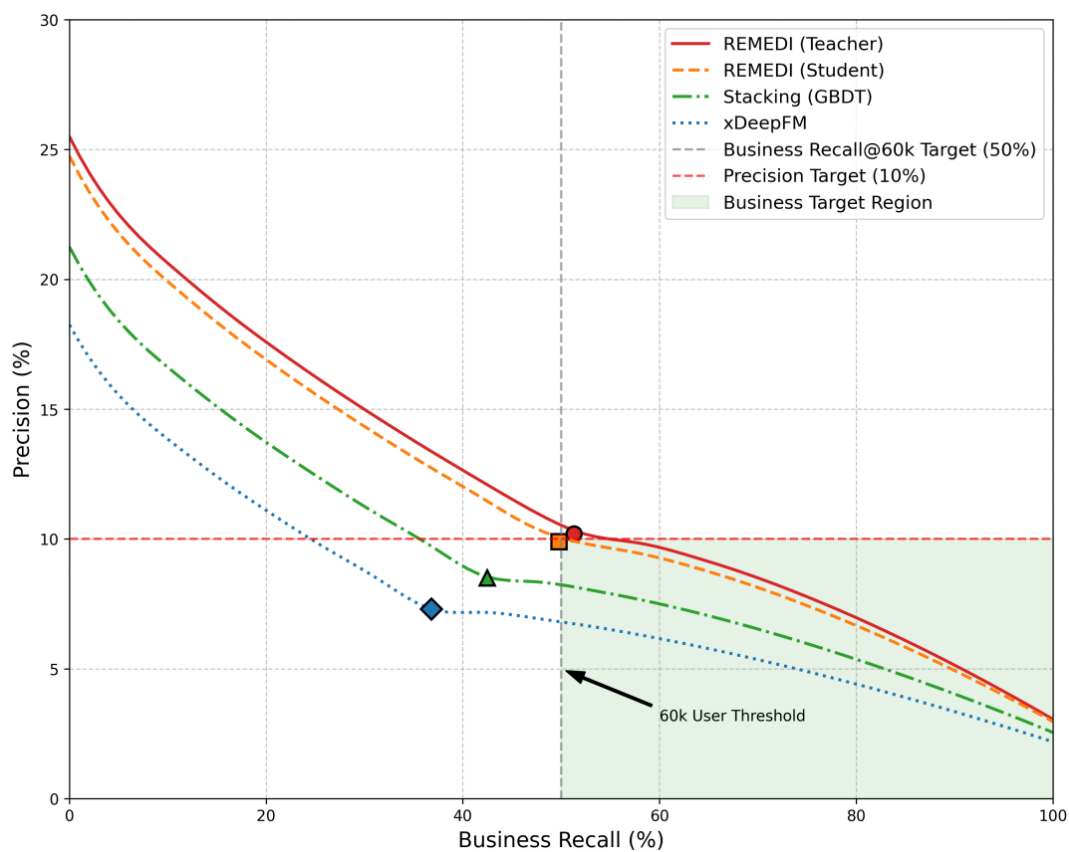

**Figure 3:** Precision-Recall Curves for Model Comparison

1. REMEDI Dominates Across Operating Points: The REMEDI teacher and student models consistently outperform baseline approaches across the entire precision-recall curve.
2. Business-Critical Region Performance: In the business-critical region around 50% recall, REMEDI maintains precision above the 10% target, while baseline approaches fall below 9%.
3. Teacher-Student Alignment: The precision-recall curves of the REMEDI teacher and student models closely track each other, confirming the effectiveness of our knowledge distillation approach.

### *4.3 Ablation Studies*

To understand individual component contributions, we conducted ablation studies by systematically modifying key elements of REMEDI.

#### *4.3.1 Impact of Meta-Learning Strategy*

We first investigated different fusion strategies while maintaining the same base models.

**Table 2:** Impact of Meta-Learning Strategy on Performance

| Fusion Strategy | Precision@60k | Business Recall@60k | AUC | Log Loss |
|---|---|---|---|---|
| Simple Average | 7.8% | 39.2% | 0.84 | 0.4 |
| Weighted Average | 8.1% | 40.6% | 0.85 | 0.39 |
| Stacking (Raw Only) | 8.5% | 42.5% | 0.87 | 0.37 |
| REMEDI (No Hybrid)* | 9.3% | 46.7% | 0.89 | 0.35 |
| REMEDI (Full) | 9.9% | 49.8% | 0.9 | 0.34 |

*"REMEDI (No Hybrid)" uses Uses relative features but with standard MLP rather than hybrid-expert architecture.

As shown in Table 2, the results demonstrate that:
1. Relative features provide substantial gains (+0.8% precision, +4.2% recall) compared to standard stacking
2. The hybrid-expert architecture adds further improvements (+0.6% precision, +3.1% recall)
3. Performance increases progressively with more sophisticated fusion strategies

*4.3.2 Importance of Base Model Diversity*

Next, we examined the impact of base model diversity through different model combinations. Table 3 illustrates that:

**Table 3:** Impact of Base Model Diversity on REMEDI Performance

| Base Model Combination | Precision@60k | Business Recall@60k | AUC |
|---|---|---|---|
| Tree Models Only (2) | 8.8% | 44.2% | 0.86 |
| Neural Models Only (2) | 8.5% | 42.6% | 0.85 |
| Mixed, No Focal Loss (4) | 9.4% | 47.1% | 0.88 |
| Mixed, No Feature-Subset (4) | 9.3% | 46.7% | 0.87 |
| All Models (5) | 9.9% | 49.8% | 0.90 |

1. Architectural diversity significantly improves performance compared to homogeneous ensembles
2. Specialized models (focal loss variant and feature-subset models) provide meaningful contributions
3. Maximum diversity with all five base models achieves the best performance.

*4.3.3 Knowledge Distillation Effectiveness*

Finally, we evaluated different distillation approaches. As presented in Table 4:

**Table 4:** Knowledge Distillation Effectiveness

| Distillation Approach | Precision@60k | Business Recall@60k | Model Size | Inference Time |
|---|---|---|---|---|
| No Distillation (Teacher) | 10.2% | 51.3% | 5+1 models | 100 ms |
| Hard Label Training* | 8.6% | 43.2% | 1 model | 10 ms |
| KL Divergence Loss | 9.7% | 48.6% | 1 model | 10 ms |
| MSE Loss (REMEDI) | 9.9% | 49.8% | 1 model | 10 ms |
| Deeper Student (MSE) | 10.0% | 50.1% | 1 model | 15 ms |

*"Hard Label Training" trains student on binary ground truth labels rather than teacher probabilities. Inference time measured per 1,000 users on standard hardware.

1. Soft probability targets significantly outperform hard label training (+1.3% precision, +6.6% recall)
2. MSE loss slightly outperforms KL divergence for knowledge transfer
3. The distilled student model achieves 97% of teacher performance with 10× speedup
4. A slightly deeper student model can further close the performance gap at minimal additional cost

*4.4 Analysis of Lead Generation*

Beyond aggregate performance metrics, we analyze the characteristics of the lead lists generated by REMEDI to gain insights into its prediction patterns and potential business impact.

*4.4.1 Score Distribution Analysis*

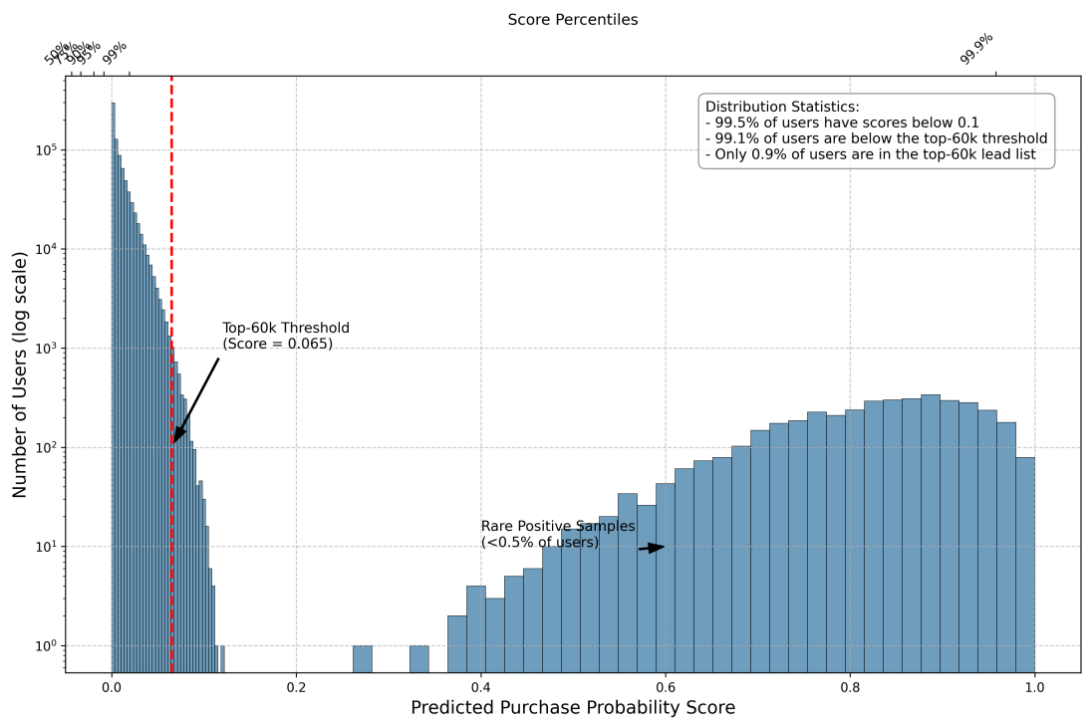

**Figure 4:** Distribution of Prediction Scores from REMEDI Student Model

Fig. 4 illustrates the distribution of prediction scores assigned by the REMEDI student model. The distribution exhibits a heavy right skew, with most users receiving very low purchase probability scores (below 0.1) and a long tail of higher-scoring users. The cutoff score for the top-60,000 users fall at approximately 0.065, representing the 92.5th percentile of the score distribution.

This pattern reflects the extreme class imbalance in the dataset, with the model assigning substantially higher probabilities to a small subset of users it confidently identifies as likely purchasers.

*4.4.2 Feature Importance Analysis*

To understand which behavioral patterns most strongly influence purchase predictions, we analyze feature importance derived from the REMEDI student model as shown in Table 5:

**Table 5:** Top 10 Features by Importance in REMEDI Student Model

| Feature | Description | Importance Score | Direction* |
|---|---|---|---|
| visit_brand_cnt | Number of unique brands visited | 13.7 | Positive |
| visit_days | Number of dealerships visit days | 11.2 | Positive |
| voice_cnt | Communication frequency related to automotive | 9.3 | Positive |
| analysis_cnt | Vertical media visit frequency | 6.0 | Positive |
| company_app_cnt | Number of manufacturer apps used | 6.0 | Positive |
| web_query_cnt_s1 | Vertical media website visit frequency | 4.7 | Positive |
| *used_request_cnt* | Second-hand vehicle app visit frequency | 4.6 | Positive |
| *voice_call_brand_cnt* | Brand diversity in communications | 3.4 | Positive |
| *visit_total_dura* | Total duration of dealership visits | 3.3 | Positive |
| *applet_flux* | Data traffic from mini-program usage | 3.0 | Positive |

*Note: "Direction" indicates whether higher feature values generally correspond to higher purchase probability.

The feature importance analysis reveals that:

1. Physical Engagement Signals Are Strong: Dealership visit-related features rank highly in importance, suggesting that physical engagement with dealerships is a powerful predictor of purchase intent.
2. Multi-Channel Behaviors Matter: The top features span multiple interaction channels, indicating that purchase intent manifests across diverse behavioral dimensions.
3. Diversity Metrics Are Particularly Predictive: Features capturing diversity of engagement show high importance, suggesting that users exploring multiple brands or engaging through multiple channels display stronger purchase signals.

## 5 Conclusion and Future Work

### 5.1 Summary of Contributions

We have presented REMEDI (Relative feature Enhanced Meta-learning with Distillation for Imbalanced prediction), a comprehensive framework addressing the challenge of predicting vehicle purchases among existing owners with extreme class imbalance (<0.5% positive rate). Our contributions include:

First, a multi-stage approach effectively combining complementary strengths of diverse prediction models while maintaining deployment practicality. By integrating heterogeneous base models with specialized training strategies, REMEDI demonstrates that architectural diversity substantially improves predictive performance for rare events.

Second, a novel meta-learning fusion mechanism inspired by comparative performance optimization, deriving relative performance features from base model predictions. This approach identifies which base models excel in specific contexts, enabling more nuanced fusion than conventional stacking.

Third, effective knowledge distillation via supervised fine-tuning with MSE loss, compressing complex ensemble knowledge into a single, deployable model while retaining 97% of performance with substantial inference efficiency gains.

Most importantly, REMEDI achieves the practical business target of ~10% precision at 50% business recall within the top-60,000 users, significantly outperforming baseline approaches. This translates to identifying approximately 6,000 likely purchasers within a manageable lead list, enabling more effective marketing resource allocation.

### 5.2 Limitations

Despite REMEDI's strong performance, several limitations merit acknowledgment:

Computational Complexity During Development: While the final distilled model is computationally efficient, the development process requires training multiple base models and a meta-model. This increased computational overhead during the development phase may pose challenges for teams with limited computational resources.

Feature Engineering Dependencies: REMEDI relies on well-engineered behavioral features that capture relevant signals across multiple interaction channels. Organizations without comprehensive behavioral tracking infrastructure may face difficulties implementing all necessary features.

Temporal Generalization: Our evaluation focuses on predicting purchases within a specific three-month window. The framework's performance for longer-term predictions remains unexplored and may require adaptation, particularly in capturing more subtle early-stage purchase indicators.

Cold-Start Scenarios: The current implementation assumes availability of behavioral history for each user. REMEDI's effectiveness for new users with limited historical data requires further investigation.

### 5.3 Future Directions

Despite REMEDI's strong performance, several limitations and future directions merit consideration:

Limitations include computational complexity during development, feature engineering dependencies requiring comprehensive behavioral tracking infrastructure, temporal generalization constraints for longer-term predictions, and potential challenges in cold-start scenarios with limited historical data.

Promising future research directions include:

- Temporal feature incorporation through sequence models (LSTM, Transformer) to capture evolving purchase intent signals
- Multi-task learning extensions for simultaneous prediction of related outcomes (purchase timing, vehicle preferences)
- Causal inference integration to distinguish predictive signals from causal factors, enhancing interpretability and intervention strategies
- Explainable AI enhancements tailored to REMEDI's architecture for improved stakeholder trust
- Adaptive retraining frameworks to detect concept drift and maintain performance stability in dynamic markets

In conclusion, REMEDI represents a significant advancement in automotive purchase prediction under extreme class imbalance. By combining diverse models through relative feature enhanced meta-learning with efficient deployment through knowledge distillation, REMEDI bridges the gap between predictive power and practical deployability, demonstrating real-world utility for optimizing marketing resource allocation.

**Acknowledgement:** The authors would like to express gratitude to the anonymous reviewers and editors for comments and suggestions

**Funding Statement:** The authors received no specific funding for this study.

**Author Contributions:** The authors confirm contribution to the paper as follows: Conceptualization, Fei Liu and Huanhuan Ren; methodology, Fei Liu and Yu Guan; software, Xiuxu Wang; validation, Fei Liu, Huanhuan Ren, and Yu Guan; formal analysis, Wang Lv; investigation, Zhiqiang Hu; resources, Yaxi Chen; data curation, Xiuxu Wang; writing—original draft preparation, Fei Liu; writing—review and editing, Huanhuan Ren and Yaxi Chen; visualization, Wang Lv; supervision, Fei Liu; project administration, Fei Liu; funding acquisition, Yaxi Chen. All authors reviewed the results and approved the final version of the manuscript.

**Availability of Data and Materials:** Data not available due to commercial and privacy restrictions. Due to the nature of this research, participants of this study did not agree for their data to be shared publicly, so supporting data is not available.

**Ethics Approval:** Not applicable.

**Conflicts of Interest:** The authors declare no conflicts of interest to report regarding the present study.

**References**

1. Choi H, Koo Y. Do I have to buy it now? A vehicle replacement model considering strategic consumer behavior. Transp Res Part D Transp Environ. 2019;73:318-337.
2. Chalil TM, Dahana WD, Baumann C. How do search ads induce and accelerate conversion? The moderating role of transaction experience and organizational type. J Bus Res. 2020;116:324-336.
3. Lin B, Shi L. Do environmental quality and policy changes affect the evolution of consumers' intentions to buy new energy vehicles. Appl Energy. 2022;310:118548. (Note: Article ID used)
4. Taylor N, Noseworthy TJ. Compensating for innovation: Extreme product incongruity encourages consumers to affirm unrelated consumption schemas. J Consum Psychol. 2020;30(1):77-95.
5. Wang Y, Gelli F, von der Weth C, Kankanhalli M. A matrix factorization based framework for fusion of physical and social sensors. IEEE Trans Multimedia. 2021;23:2782-2793.
6. Yan Q, Zhang Y, Liu Q, Wu S, Wang L. Relation-aware heterogeneous graph for user profiling. In: Proceedings of the 30th ACM International Conference on Information & Knowledge Management (CIKM '21). 2021. p. 2376-2386. (Note: Conference Paper formatted similarly to journal)
7. Behera RK, Gunasekaran A, Gupta S, Kamboj S, Bala PK. Personalized digital marketing recommender engine. J Retail Consum Serv. 2019;53:101799. (Note: Article ID used)
8. Zhang W, Bao W, Liu XY, Yang K, Lin Q, Wen H, et al. Large-scale causal approaches to debiasing post-click conversion rate estimation with multi-task learning. In: Proceedings of The Web Conference 2020 (WWW '20). 2020. p. 1737-1747. (Note: Conference Paper formatted similarly to journal)
9. Puy A, Beneventano P, Levin SA, Lo Piano S, Portaluri T, Saltelli A. Models with higher effective dimensions tend to produce more uncertain estimates. Sci Adv. 2022;8(42):eabn9450. (Note: Article ID used)
10. Ozdemir A. Development of a bi-objective 0-1 mixed-integer nonlinear response surface-based robust design optimization model for unbalanced experimental data. Comput Ind Eng. 2021;158:107446. (Note: Article ID used)
11. Rezvani S, Wang X. A broad review on class imbalance learning techniques. Appl Soft Comput. 2023;143:110418. (Note: Article ID used)
12. Dina AS, Siddique AB, Manivannan D. A deep learning approach for intrusion detection in internet of things using focal loss function. Internet Things. 2023;22:100699. (Note: Article ID used)
13. Serban A, Visser J. Adapting software architectures to machine learning challenges. In: 2022 IEEE International Conference on Software Analysis, Evolution, and Reengineering (SANER). 2022. p. 152-163. (Note: Conference Paper formatted similarly to journal)
14. Lv SX, Peng L, Hu H, Wang L. Effective machine learning model combination based on selective ensemble strategy for time series forecasting. Inf Sci. 2022;612:994-1023.
15. Hassan MF, Abdel-Qader I, Bazuin B. A new method for ensemble combination based on adaptive decision making. Knowl Based Syst. 2021;233:107544. (Note: Article ID used)
16. Paleyes A, Urma RG, Lawrence ND. Challenges in deploying machine learning: a survey of case studies. ACM Comput Surv. 2023;55(6):Article 138. (Note: Article ID/format)
17. Huang Y, Wang H, Miao Y, Xu R, Zhang L, Zhang W. Neural statistics for click-through rate prediction. In: Proceedings of the 45th International ACM SIGIR Conference on Research and Development in Information Retrieval (SIGIR '22). 2022. p. 1054-1063. (Note: Conference Paper formatted similarly to journal)
18. Liu M, Cai S, Lai Z, Qiu L, Hu Z, Ding Y. A joint learning model for click-through prediction in display advertising. Neurocomputing. 2021;445:206-219.
19. Chen Y, Wang Y, Ren P, Wang M, de Rijke M. Bayesian feature interaction selection for factorization machines. Artif Intell. 2021;302:103589. (Note: Article ID used)
20. Fei L, Lili L, Kai H, Teng C, Suxia Z. Movie recommendation system for educational purposes based on field-aware factorization machine. Mob Netw Appl. 2021;26(5):2199-2205.
21. Sagi O, Rokach L. Approximating XGBoost with an interpretable decision tree. Inf Sci. 2021;572:522-542.
22. Wang DN, Li L, Zhao D. Corporate finance risk prediction based on LightGBM. Inf Sci. 2022;602:259-268.
23. Tang B, Tang M, Xia Y, Hsieh MY. Composition pattern-aware web service recommendation based on depth factorisation machine. Connect Sci. 2021;33(4):870-890.
24. Lu Q, Li S, Yang T, Xu C. An adaptive hybrid XdeepFM based deep interest network model for click-through rate prediction system. PeerJ Comput Sci. 2021;7:e716. (Note: Article ID used)
25. Gan M, Ma Y. DeepInteract: Multi-view features interactive learning for sequential recommendation. Expert Syst Appl. 2022;204:117305. (Note: Article ID used)

18    J Artif Intell. 2025;volume(issue)

26. Wang R, Shivanna R, Cheng D, Jain S, Lin D, Hong L, et al. DCN V2: Improved deep & cross network and practical lessons for web-scale learning to rank systems. In: Proceedings of The Web Conference 2020 (WWW '20). 2020. p. 1785-1795. (Note: Conference Paper formatted similarly to journal)
27. Kim B, Ko Y, Seo J. Novel regularization method for the class imbalance problem. Expert Syst Appl. 2021;188:115974. (Note: Article ID used)
28. Chen Q, Zhang ZL, Huang WP, Wu J, Luo XG. PF-SMOTE: A novel parameter-free SMOTE for imbalanced datasets. Neurocomputing. 2022;498:75-88.
29. Wang Z, Cao C, Zhu Y. Entropy and confidence-based undersampling boosting random forests for imbalanced problems. IEEE Trans Neural Netw Learn Syst. 2020;31(12):5178-5191.
30. Fernando KRM, Tsokos CP. Dynamically weighted balanced loss: Class imbalanced learning and confidence calibration of deep neural networks. IEEE Trans Neural Netw Learn Syst. 2021;33(7):2940-2951.
31. Jiang F, Yu X, Du J, Gong D, Zhang Y, Peng Y. Ensemble learning based on approximate reducts and bootstrap sampling. Inf Sci. 2021;547:797-813.
32. Kadkhodaei HR, Eftekhari Moghadam AM, Dehghan M. HBoost: A heterogeneous ensemble classifier based on the boosting method and entropy measurement. Expert Syst Appl. 2020;157:113482. (Note: Article ID used)
33. Kshatri SS, Singh D, Narain B, Bhatia S, Quasim MT, Sinha GR. An empirical analysis of machine learning algorithms for crime prediction using stacked generalization: an ensemble approach. IEEE Access. 2021;9:67488-67500.
34. Tian Y, Chen M, Shen L, Jiang B, Li Z. Knowledge distillation with multi-objective divergence learning. IEEE Signal Process Lett. 2021;28:962-966.
35. DeepSeek-AI, Guo D, Yang D, Zhang H, Song J, Zhang R, et al. DeepSeek-R1: Incentivizing Reasoning Capability in LLMs Via Reinforcement Learning. CoRR 2025; abs/2501.12948.
36. Fu Z, Li X, Wu C, Wang Y, Dong K, Zhao X, et al. A Unified Framework for Multi-Domain CTR Prediction Via Large Language Models. arXiv [Preprint] 2023; abs/2312.10743.
37. Zhang CB, Jiang PT, Hou Q, Wei Y, Han Q, Li Z, et al. Delving deep into label smoothing. IEEE Trans Image Process. 2021;30:5984-5996.
38. Xi Y, Liu W, Lin J, Cai X, Hong Z, Zhu J, et al. Towards Open-World Recommendation with Knowledge Augmentation from Large Language Models. Computing Research Repository 2024; 12-22.